\documentclass[conference,final]{IEEEtran}
\IEEEoverridecommandlockouts
\usepackage{amsmath,amssymb,amsfonts}
\usepackage{graphicx}
\usepackage{textcomp}
\usepackage{xcolor}
\usepackage{grffile}
\usepackage{tikz}
\usepackage{float}
\usepackage{nccmath}
\usepackage{amsfonts}
\usepackage{amsmath}
\usepackage{subcaption}
\usepackage{geometry}
\usepackage{multirow}
\usepackage{booktabs}
\usepackage{lipsum}
\usepackage{caption}
\usepackage{nicefrac}
\usepackage{mathtools}
\usepackage{siunitx}
\usepackage{ulem}
\usepackage{listings}
\usepackage{color,soul}
\usepackage{fancyvrb}
\usepackage{hyperref}
\hypersetup{ 
    colorlinks=true,
    linkcolor=blue,
    filecolor=magenta,      
    urlcolor=cyan,
}

\newcolumntype{d}[1]{D{.}{.}{#1}}

\makeatletter
\newcommand\footnoteref[1]{\protected@xdef\@thefnmark{\ref{#1}}\@footnotemark}
\makeatother

\newcommand*\circled[1]{\tikz[baseline=(char.base)]{\node[shape=circle,draw,inner sep=0.25pt] (char) {\footnotesize{#1}};}}
\newcommand{\X}{\phantom{0}}

\newtheorem{definition}{Definition}
\DeclareMathOperator{\median}{median}
\newcommand{\p}{\ensuremath \mathbf{p}}

\newcommand\copyrighttext{%
  \footnotesize \textcopyright 2021 IEEE. Personal use of this material is permitted.
  Permission from IEEE must be obtained for all other uses, in any current or future 
  media, including reprinting/republishing this material for advertising or promotional 
  purposes, creating new collective works, for resale or redistribution to servers 
  or lists, or reuse of any copyrighted component of this work in other works. 
}

\def\BibTeX{{\rm B\kern-.05em{\sc i\kern-.025em b}\kern-.08em
    T\kern-.1667em\lower.7ex\hbox{E}\kern-.125emX}}
\begin{document}


\title{Detecting Wandering Behavior of People with Dementia \\ 
\thanks{We acknowledge financial support from TrygFonden and the Danish municipality of Nyborg for the project "Sammen Om Demens" (\url{https://sod.sdu.dk}), to which this work belongs.}
}

\author{\IEEEauthorblockN{Nicklas Sindlev Andersen, Marco Chiarandini, Stefan J\"anicke, Panagiotis Tampakis, Arthur Zimek}
\IEEEauthorblockA{\textit{Institute of Mathematics and Computer Science} \\
\textit{University of Southern Denmark}\\
Odense, Denmark \\
\{sindlev$|$marco$|$stefan$|$ptampakis$|$zimek\}@imada.sdu.dk}
}

\maketitle

\begin{tikzpicture}[remember picture,overlay]
  \node[anchor=south,yshift=10pt,xshift=-15pt] at (current page.south) {\fbox{\parbox{\dimexpr0.998\textwidth-\fboxsep-\fboxrule\relax}{\copyrighttext}}};
\end{tikzpicture}%

\begin{abstract}
  Wandering is a problematic behavior in people with dementia that can lead to dangerous situations.
  To alleviate this problem we design an approach for the real-time automatic detection of wandering leading to getting lost. The approach relies on GPS data to determine frequent locations between which movement occurs
  and a step that transforms GPS data into geohash sequences. Those can be used to find frequent and normal movement patterns in historical data to then be able to determine whether a new on-going sequence is anomalous. We conduct experiments on synthetic data to test the ability of the approach to find frequent locations
  and to compare it against an alternative, state-of-the-art approach. Our approach is able to identify frequent locations and to obtain good performance (up to AUC = 0.99 for certain parameter settings) outperforming the state-of-the-art approach.  
\end{abstract}


\normalem 

\section{Introduction} \label{sec:introduction}

The number of persons affected by dementia is expected to increase worldwide \cite{war2015}. Persons affected by dementia are known to get disoriented causing them to wander and get lost. This is a concern because a person can face dangerous situations, which provokes anxiety in relatives and care-taking persons. Our work has been motivated by a collaboration with social workers who aim at supplying people with dementia with a smartphone app that notifies relatives when a person starts to get disoriented causing them to stray from their normal travel patterns.

There have been proposed several approaches on how disorientation and wandering can be defined in persons suffering from dementia~\cite{LIN201449, Has2019, Mar1991}. Many of the approaches that are based on the Martino-Saltzman model~\cite{Mar1991}, successfully detect wandering patterns but they do not take into consideration the historical travel patterns of a person, which might vary from individual to individual. 
Other approaches are of a more personalized nature and focus on a definition of wandering as a deviation from the normal patterns of a person~\cite{Spo2010,Yin2008,Cha2010,LIN2015,WOJ2019}.

\begin{figure}[t]
    \centering
    \includegraphics[trim={0.5cm 0.5cm 0.5cm 0.90cm},clip, width=0.5\textwidth]{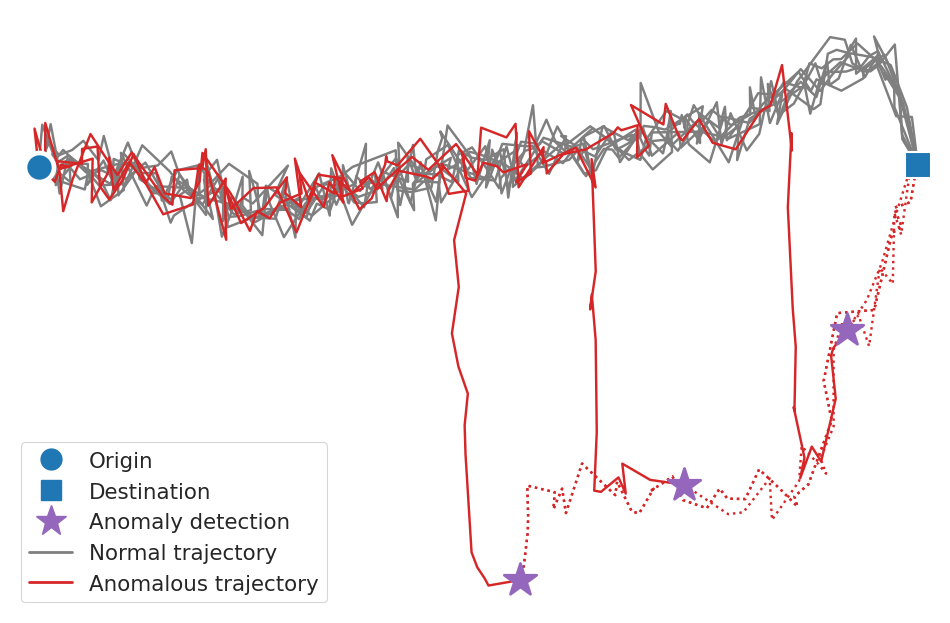}
    \caption{An illustrative example of normal and anomalous trajectories going between a single origin and destination. Trajectories deviating from the normal movement pattern have been detected and labeled as being anomalous.}
    \label{fig:intro}
\end{figure}

\begin{figure*}[tb!]
    \centering
    \includegraphics[trim={2.75cm 9.80cm 4.50cm 1.0cm}, clip, width=1\textwidth]{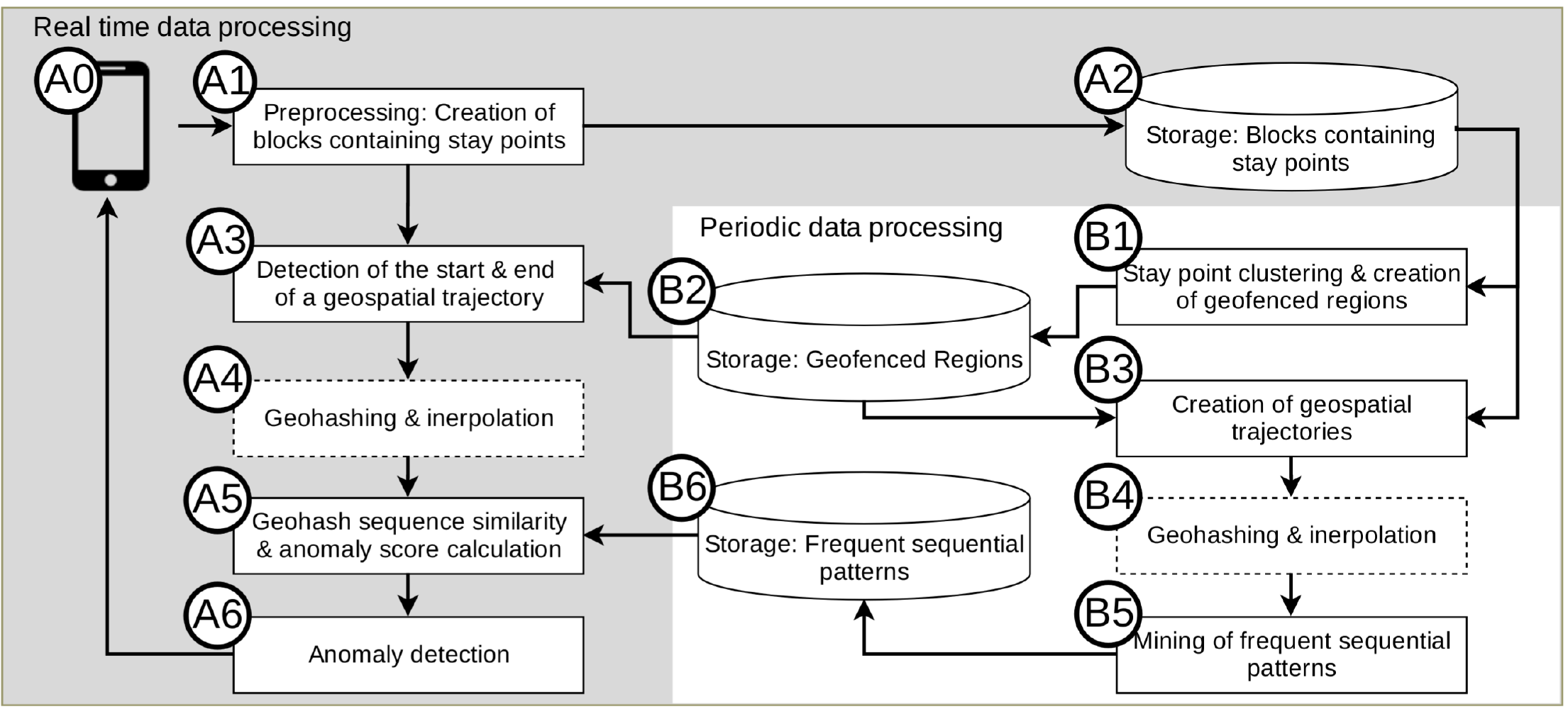}
    \caption{An overview of the components of the approach displaying their organization and the data processing flow.}
    \label{fig:data_processing_flow}
\end{figure*}

In this paper, we follow the latter approach and propose an algorithm for the real-time automatic detection of disorientation behavior from GPS data. Fig.~\ref{fig:intro} shows an example of the input and output of our algorithm. Given a set of historical paths between an origin and a destination determined by the locations transmitted, the algorithm is able to detect in real-time an incumbent path as anomalous if it deviates considerably from the normal pattern. When such behavior is detected, relatives and nearby volunteers can be automatically alerted and called to provide assistance. In this paper, we describe the algorithm and also the complete data processing pipeline from data receipt and preprocessing to the final detection result. This pipeline is depicted in Fig.~\ref{fig:data_processing_flow}.

The most similar approach to ours by Lin et al.~\cite{LIN2015} represents the movement of elderly people as sequences of ``stops'' that correspond to semantic places (e.g., home, grocery shop, etc.) and ``moves'' that correspond to the movement of a person between ``stops''. The algorithm identifies anomalous movement by examining whether the movement of an elderly person deviates significantly from historical sequences of movement.
However, the authors assume that semantic places are given a priori while our problem setting requires their discovery in an online manner while data are streaming. 
Furthermore, Lin et al. discretize the movement of elderly people by using a uniform grid, which can be quite inefficient, while we use geohashing that is a more flexible and distance-preserving way of discretizing movement into sequences.
Moreover, Lin et al.'s algorithm~\cite{LIN2015} tries to match the entire real-time sequence of movement of a person exactly with a subsequence of a historical sequence of movement of the same person. Clearly, this might be too restrictive in a context where location positions might contain noise and trajectories may deviate slightly without being anomalous. In contrast, we adopt a similarity function~\cite{SMI81} between real-time and historical sequences, which is more flexible and allows for small deviations between sequences. 
Consequently, while in Lin et al.'s algorithm~\cite{LIN2015} all historical sequences of a person must be compared with the real-time sequence, we can apply a frequent pattern mining algorithm periodically, store only the frequent patterns and compare the real-time sequence only with those frequent patterns.

The remainder of this paper is structured as follows. 
In Section~\ref{sec:preliminaries}, we describe the components of the proposed disorientation detection approach and formalize the problem. In Section~\ref{sec:experimental_setup}, we describe the experimental setup used in the evaluation of the proposed detection approach along with the synthetically generated data used in the proceeding algorithm experiments. The results from the experiments with the proposed detection approach and the alternative~\cite{LIN2015} are then given in Section~\ref{sec:experimental_results}. Finally, in Section~\ref{sec:conclusion} we draw conclusions and outline future work.

\section{Data Processing \& Detection Task} \label{sec:preliminaries}

The components of the wandering detection approach visualized in Fig.~\ref{fig:data_processing_flow} can be divided into two groups: \circled{A0}--\circled{A6} are components that process data in real-time, while \circled{B1}--\circled{B6} are components that run periodically and use historical data to support the functioning of components \circled{A3} and \circled{A5}.

\subsection{Data preprocessing}
\label{sub:preliminaries}

Raw GPS data are processed in \circled{A1} in real-time as they arrive from a smartphone device \circled{A0}. Raw GPS data come in the form of \emph{geospatial points} consisting of a vector $\p=[x,y] \in \mathbb{R}^2$ where $x$ and $y$ are the latitude and longitude coordinate, respectively, in any geographical coordinate system. 

\begin{definition}[Stream of geospatial points]
  A stream of geospatial points is a sequence of ordered pairs $S = [(\p_{i}, \delta_{i})]_{i=1, \dots, n}$ where $\p_{i}$ is a geospatial point and $\delta_{i}$ is a timestamp.
\end{definition}

We use the Haversine distance $d: \mathbb{R}^2\times \mathbb{R}^2\to \mathbb{R}$ to approximate distances between points on Earth. The stream of geospatial points incorporates a temporal dimension in which the relation between consecutive points can be assessed. The stream of geospatial points can be organized into \emph{blocks} based on the temporal and spatial relation between consecutive points.

\begin{definition}[Block]
    A stream of points can be partitioned in a set of blocks $\{B_{i}\}_{i=1, \dots, m}$ starting and ending at two consecutive points that are at a distance in time larger than $\epsilon \in \mathbb{R}_+$ or in space larger than $\gamma \in \mathbb{R}_+$, i.e., a block is a sequence of consecutive geospatial points from a stream $S$ starting at any $\ell=1, \dots, n$ element such that $\delta_{\ell} - \delta_{\ell-1} \geq \epsilon$ or $d(\p_{\ell-1}, \p_{\ell}) \geq \gamma$ and ending at the data point before the successive block starts.
\end{definition}

Blocks organize a stream of points into meaningful groups, in the sense that if consecutive points are far apart in space and time then it can be assumed that their relation is insignificant, i.e., they are not part of the same trip that a person makes. Large temporal and spatial gaps between points can for example be due to the GPS sensor being deliberately switched off by the person or due to bad satellite coverage causing the GPS sensor to be idle.

The parameters $\epsilon$ and $\gamma$ are distance thresholds that specify upper bounds on the amount of information we are prepared to infer about the movement of a person between two consecutive geospatial points. Setting the thresholds too high introduces uncertainty about the movement of a person between consecutive points, while too small values will make it impossible to capture the larger spatial and temporal context of the movement of a person. 

A \emph{stay point} $\p'_{i}$ in a block $B$ is a geospatial point augmented with a weight $w'_i \in \mathbb{R}_+$, which is a duration of time, that results from the contraction of all consecutive points from $i$ to $j$, $[(\p_{\ell}, \delta_{\ell})]_{\ell=i, \dots, j}$, and that is defined with respect to a tentative point 
\begin{align} \label{contraction1}
    \medmath{\tilde{\p}_{ij} = (x'_{i}, y'_{i}) = (\median(x_{i}, \dots, x_{j}), \median(y_{i}, \dots, y_{j}))}
\end{align}

A stay point satisfies the conditions $d(\tilde{\p}_{ik}, \p_{k + 1}) < \xi'$ for all $k = i, \dots, j - 1$, and $\max_{\ell = i, \dots, k}{d(\tilde{\p}_{ik}, \p_{\ell})} < \alpha \in \mathbb{R}_+$ for all $k = i, \dots, j$, and replaces $[(\p_{\ell}, \delta_{\ell})]_{\ell=i, \dots, j}$ with
\begin{align} \label{contraction2}
    &\medmath{
        \p'_{i} = \tilde{\p}_{ij}
    }, \;
    \medmath{
        \delta_{i} = \delta_{j}
    },\; 
    \medmath{
        w'_{i} = \delta_{j} - \delta_{i}
    }\;
\end{align}

The process of forming stay points in a block can perhaps be more easily understood as a data compression task performed online. Every time a new data point $\p_{i + k}$ arrives, it is checked whether (a) its distance from the median of the $i$ to $i + k-1$ previous points is smaller than $\xi'$, and (b) if a new median of the $i$ to $i + k$ points will still be within a distance $\alpha$ of each of the $i$ to $i+k$ points. If conditions (a) and (b) are met then the points are to be contracted together. However, the contraction operation is suspended until a point $\p_{j+1}$ arrives that is either farther away than $\xi'$ from the median of the $i$ to $j$ previous points or the median of the $i$ to $j+1$ points are farther away than $\alpha$ distance from each of the $i$ to $j+1$ points. Then, the points $\p_i,\dots,\p_j$ are contracted into $\p'_i$ as described by Eqs.~{\eqref{contraction1}}-{\eqref{contraction2}}. The primary goal of this process is to reduce the amount of data that needs to be processed in a subsequent step. Secondarily, the process aims to (i) obtain more robust geospatial coordinates, and to (ii) assign appropriate weights to locations where a person remains stationary for an extended period of time.

The value of $\xi'$ should be chosen by considering the inaccuracy of a smartphone GPS sensor and the level of data compression that should be achieved. Noisy geospatial points that are scattered around the true location of a person should preferably be replaced by a stay point to benefit from (i) and (ii). The inaccuracy of a smartphone GPS sensor has been reported in {\cite{Kri19}} to be in the range of $7$-$13m$ on average. Furthermore, the condition $\xi' < \gamma$ should be satisfied to avoid conflicts with the subdivision of a stream of points into blocks. Taking these things into account, a suitable value for $\xi'$ would lie in the range $13 < \xi' < \gamma$.

The distance threshold $\alpha$ is introduced to ensure the possible drift of a stay point is confined to be within a certain distance of the geospatial points that are used to form the stay point. The drift of a stay point can for example be due to a too large value of $\xi'$ or because a person is walking very slowly between two locations. As a guideline, the value of $\alpha$ should lie in the range $\xi' < \alpha < \gamma$.

The preprocessing step \circled{A1} ultimately creates blocks containing stay points that are saved in \circled{A2} to be used in the future but are also further processed in real-time \circled{A3}--\circled{A6}.

\subsection{Subdivision of blocks in trajectories} \label{sub:clustering}

Let $X(B)$ denote a block of data after the contraction in stay points. Points that are not contracted are also augmented with a weight equal to zero and become stay points as well.
Once enough blocks have been collected and processed in this way in \circled{A2}, clustering of stay points from different blocks of a person is performed in \circled{B1}.

The clustering works as follows.
Let the stream $S$ contain $m$ blocks and let the blocks be contracted, i.e.~$\{X(B_{i})\}_{i=1, \dots, m}$.
Stay points with a weight $w > \tau \in \mathbb{R}_+$ are considered as nodes in a network and all pairs of nodes in the network, that are within some distance $\xi'' \in \mathbb{R}_+$ of each other, are connected with an edge. Each edge is given a weight equal to the inverse normalized distance between the pair of nodes. This serves the purpose of relating stay points, that can be found across different blocks, with each other. The grouping of the stay points is found by applying a network clustering algorithm: the Louvain community detection algorithm \cite{BLO08}. The resulting clusters are used to form \emph{geofenced regions}.

The parameter $\tau$ is used as a threshold for pruning uninteresting stay points that might be the result of the slow movement of a person or minor stops at e.g. an intersection. The parameter $\xi''$ has a somewhat similar physical interpretation as $\xi'$ and is used as a measure for deciding when one and several other stay points in a general area is thought to be highly related to each other while taking into account the fact that stay points from different blocks might be scattered around a general area due to inherent inaccuracies of the data.

Appropriate values for $\tau$ and $\xi''$ depend heavily on the value chosen for the parameter $\xi'$, as $\xi'$ essentially decides when and which geospatial points are replaced with a stay point. This in turn has a direct effect on the geospatial coordinates and weight given to a stay point. Due to the interaction of the parameters, the values for $\tau$ and $\xi''$ are thus best decided through experimentation.

\begin{definition}[Geofenced region]
  A geofenced region is a geospatial region defined by the convex hull of all stay points $\p_{i}$ belonging to an identified cluster.
\end{definition}

The geofenced regions represent meaningful geospatial regions that a person frequently visits, e.g., a person's home, a grocery store, etc. The regions identify the natural start and end points for subdividing a block in real-time in \circled{A3}, and historically collected blocks in \circled{B3}, into \emph{geospatial trajectories}.

\begin{definition}[Geospatial trajectory]
  A geospatial trajectory is a subset of a contracted block of stream data delimited by the geofenced regions, that is, $T = \{\p_{0}, \dots, \p_{i}, \dots, \p_{n}\}$ where $\p_{0}$ and $\p_{n}$ belong to the same or two different geofenced regions and all other points $\p_i$ for $i=1,\ldots,n-1$ do not.
\end{definition}

A geospatial trajectory corresponds to the movement of a person between geofenced regions.

\subsection{Geohashing}

\begin{figure}[t]
  \centering
  \includegraphics[trim={1.25cm 14.5cm 24cm 0.cm},clip,width=0.5\textwidth]{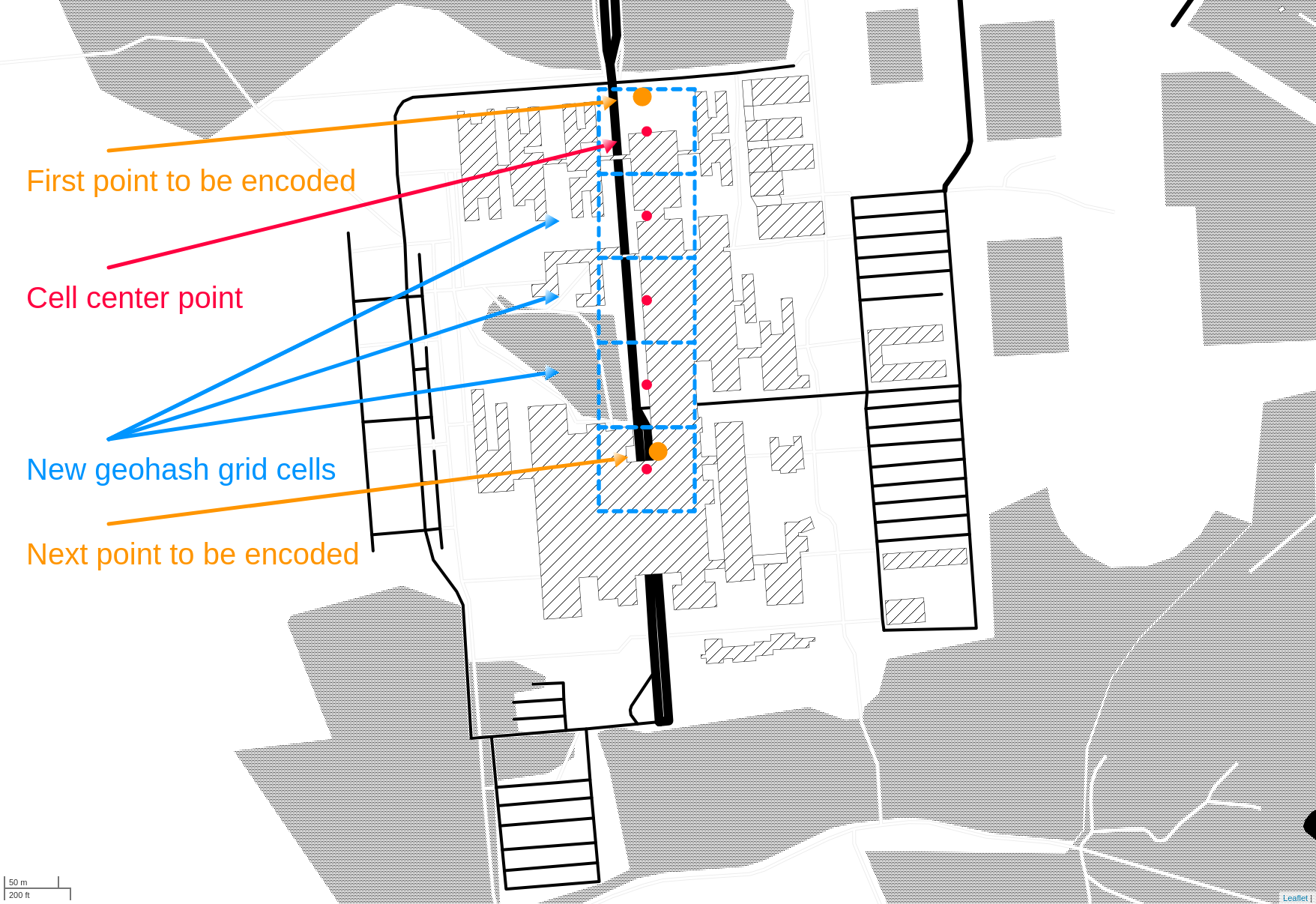}
  \caption{Illustration of the geohash sequence interpolation process.}
  \label{fig:interpolation_process}
\end{figure}

In \circled{A4} and \circled{B4}, the coordinates of stay points forming a geospatial trajectory are transformed into their corresponding geohash, forming \emph{sequences of geohashes}. An interpolation process enriches geohash sequences by adding missing cells between geohash grid cells if these are not connected. Furthermore, consecutive and identical geohash grid cells are merged. The result is illustrated in Fig~\ref{fig:interpolation_process}. Note that the contractions performed at the previous stages reduce the number of points that need to be hashed and considered in the next computations thus speeding up the processing. Note also that the geohash sequences will have the first element contained in a geofenced region. The next element will be the first outside of the geofenced region.

\subsection{Sequential pattern mining} \label{sub:sequential_pattern_mining}

Offline, the data in \circled{A2} are processed in a set $H$ of historical geohash sequences. In \circled{B5}, we extract from $H$ a set $D$ of \emph{frequent sequential patterns}
that represents the normal movement patterns of a person. The set $D$ is generated by mining frequent sequential patterns via the PrefixSpan algorithm~\cite{PEI2001}. The PrefixSpan algorithm finds sequence patterns that occur in a database of sequences. A frequent sequential pattern is \emph{closed} if no super-pattern with the same frequency count exists. Furthermore, patterns that are contained in a super-pattern, irrespective of frequency count, are termed redundant patterns and pruned from the resulting set. The goal of this step is to extract the most important information from $H$ while reducing the overall set of sequences to be saved in \circled{B6} and to be considered in the subsequent sequence similarity calculation step \circled{A5}.
Thus, at each mining step we only store in $D$ the closed sequence patterns with a frequency higher than $\eta$ and that has been pruned, as illustrated in the example in Fig.~\ref{prefix}.

\begin{figure}[t]
\begin{minipage}{.495\columnwidth}
\begin{lstlisting}
H = [["A", "B", "D"],
    ["A", "B", "C", "D"],
    ["A", "B", "C", "D"],
    ["A", "B", "E", "F"],
    ["A", "B", "E", "F"],
    ["A", "B", "E", "F"],
    ["A", "E", "F"],
    ["A", "G", "H", "I"],
    ["A", "G", "H", "I"]]
\end{lstlisting}
\end{minipage}
\begin{minipage}{.495\columnwidth}
\begin{lstlisting}
[
$\times$ |(9, ['A']),|
$\times$ |(6, ['A', 'B']),|
$\phantom{\times}$ (3, ['A', 'B', 'D']),
$\phantom{\times}$ (2, ['A', 'B', 'C', 'D']),
$\phantom{\times}$ (3, ['A', 'B', 'E', 'F']),
$\phantom{\times}$ (4, ['A', 'E', 'F']),
$\phantom{\times}$ (2, ['A', 'G', 'H', 'I']),
]
\end{lstlisting}
\end{minipage}
\caption{\label{prefix} The result of the PrefixSpan algorithm (right) when applied to database $H$ containing sequences between geofenced regions \texttt{A}-\texttt{D}, \texttt{A}-\texttt{F} and \texttt{A}-\texttt{I} (left) restricted to closed patterns that have been pruned ($\times$) and $\eta = 2$.}
\end{figure}

The operations in \circled{B1}, \circled{B3}, \circled{B4}, \circled{B5} are repeated in the background periodically as with new data arriving in \circled{A2} the outcome $D$ may change.

\subsection{Sequence similarity calculation}

In \circled{A5}, the set of frequent sequential patterns belonging to a person is used for calculating a \emph{similarity score} of an on-going geohash sequence generated by the same person. 

The similarity score is a distance measure used to quantify the similarity between two sequences. Different similarity measures between sequences are possible. Some distance measures are able to handle sequences of different lengths, while others like the Hamming distance only applies to sequences of the same length. Some distance measures do not take into consideration the ordering of the elements in a sequence. For example, the Jaccard similarity can measure the degree of overlap between two sets of sequences, but without considering the ordering of the elements.

The edit distance is one of the most commonly used distance measures for comparing sequences. It counts the minimum number of operations required to transform one sequence into the other. The edit distance is a generalization of other well-known sequence similarity measures such as the Hamming distance, Longest Common Subsequence (LCSS), and the Levenshtein distance. The only difference among these distances is the set of allowed operations (insertion, deletion, or substitution of elements) that can be used to transform one sequence into the other.

A further generalization of the edit distance comes in the form of sequence alignment algorithms where each allowed operation is assigned a cost and the cost of the operation depends on the location where it is applied. The pairwise Smith-Waterman sequence alignment algorithm is an example of such an algorithm \cite{SMI81}. The Smith-Waterman algorithm is suitable for comparing partially similar sequences and sequences of different lengths that have conserved regions with high similarity. It is one of the most commonly used sequence alignment algorithms.

We adopt this algorithm to calculate the similarity score. The on-going sequence of a person is compared in turn to each sequence in the set of frequent sequential patterns $D$. The best (largest positive) similarity score found among all the pairwise comparisons is then used to assess how anomalous the on-going sequence of the person is. 

More precisely, the similarity score is calculated by the following function that takes as input (i) a set of frequent sequential patterns $D$ and (ii) an on-going sequence $V$ being generated by a person: 

\begin{align} \label{eq:simcalc}
  F(D, V) = \max_{\forall\,U \in D} \
    \left(
        \nicefrac{
            f_{\mathrm{sw}}\left(U, V\right)
        }{
            |V|
        }
    \right)
\end{align}
where $f_\mathrm{sw}$ represents the application of the pairwise Smith-Waterman sequence alignment algorithm and counts how many elements of the sequence $U$ can be matched with the elements of the sequence $V$. The function $F$ thus takes values in the interval $[0, 1]$. A value of $1$ means that there is a perfect match between the on-going sequence $V$ and a sequence pattern $U$ in $D$. On the other hand, a value of $0$ means there is a complete mismatch between the on-going sequence and the patterns in $D$.

Every time a new element $v_i$ is added to an on-going geohash sequence $V_i=[v_0,v_1,\ldots,v_{i-1},v_i]$ in \circled{A4}, the similarity score is recalculated $s_i = F(D, V_i)$ and an anomaly score $a_i$ associated to $V_i$ is updated as follows: $a_i = 1 - \min_{j = 0,\dots, i}\left(s_j\right)$.

\subsection{Anomaly Detection}

We are now ready to formalize the anomaly detection task as follows:
Given a database $D$ of frequent sequential patterns $D$ and an incoming block $B$, the goal is to detect whether the last trajectory of the block transformed in a geohash sequence is anomalous with respect to $D$. 

We address this task in \circled{A6} by monitoring the anomaly score. If its value raises above some anomaly threshold $\theta \in (0, 1] $, then the trajectory is classified as anomalous and an alarm procedure is triggered.

\section{Experimental Setup} \label{sec:experimental_setup}

We use a machine with an AMD Ryzen 7 1800X eight-core processor and 32 GB of RAM, running Manjaro Linux, to execute the experiments. The implementation\footnote{\label{git_repo} \url{https://github.com/nicklasxyz/rtdm}} was done in Python primarily utilizing the libraries: Biopython \cite{biopython}, Geohash-hilbert \cite{geohash-hilbert}, Pandas \cite{mck2010},  PrefixSpan-py \cite{prefixspan-py} and Scikit-learn \cite{sklearn}.

\subsection{Case studies} \label{sub:case_studies}

We evaluate the proposed approach and compare it with the iBDD method \cite{LIN2015}. We use synthetically generated data\footnoteref{git_repo} since (i) it is possible to specifically generate anomalous trajectories with the target characteristics given in Section~\ref{sec:introduction}, and (ii) the trajectories can automatically be labeled according to what they were generated as (normal or anomalous). Because of (ii) it becomes straightforward then to quantitatively evaluate the results. More precisely, the approach presented can be assessed in a supervised setting, that is, statistics can be computed based on whether a trajectory has been labeled correctly by the approach or not (detected as correctly/incorrectly being normal or anomalous). We considered other open-source and real-life trajectory data sets, such as e.g.~the \emph{GeoLife Trajectory Dataset} released by Microsoft research Asia\footnote{ \;\url{https://www.microsoft.com/en-us/download/details.aspx?id=52367}}, but in this case, it is unknown whether the data explicitly contain travel patterns that result from disorientation behavior. 

We consider two case-studies: (1) a case-study used for showing the results of the preprocessing and segmentation steps described in Section~\ref{sub:preliminaries} and Section~\ref{sub:clustering}, and (2) a large case-study used for an evaluation and comparison of the proposed approach against the iBDD method that maintains a set of historical trajectories that has been turned into a corresponding set of historical sequences of traversed cells (dubbed a \emph{support set}). Based on the set of historical sequences, the iBDD method labels a new on-going sequence as being anomalous if less than a certain fraction of the historical sequences support the current on-going sequence. That is, like our approach the iBDD relies on two parameters: a parameter for determining the geohash precision, and a threshold $\theta'\in (0, 1]$ that is the fraction of historical sequences needed to support an on-going sequence before it is otherwise considered anomalous. The iBDD method also relies on a parameter that specifies the rate at which new geospatial points arrive. This parameter is mainly used for the purpose of filtering out noisy points and can simply be estimated from historical data.

Case-study (1) consists of a single synthetically generated, normal trajectory going back and forth between an origin and a destination, simulating a person walking between two locations and staying at each location for an extended period of time (15 min).

Case-study (2) consists of 222 synthetically generated normal and anomalous trajectories going between 7 different origins and destinations resulting in a variety of different trajectories. The trajectories generated as being normal are 199 while those being anomalous are 23. In other terms, around 90\% of the trajectories are normal and 10\% are anomalous.

\subsection{Synthetic data}
 
The approach used for generating the synthetic trajectories is a slight variation of the approach used by Hermoupolis \cite{Pel15}, a semantic trajectory generator. Hermoupolis is a network-based trajectory generator that is able to generate trajectories for several moving entities at a time. The trajectories accurately follow an underlying road network and take into account the effects of network constraints.

We generate trajectories for a single person mimicking the characteristics of trajectories that arise from GPS data resulting from walking. In other words, an underlying pedestrian road network and normal pedestrian walking speed (approximately $4.5 \frac{m}{s}$) are used in the generation. 

Normal trajectories are generated by simulating, one at a time, an entity that moves along an underlying path in a given road network between a specified geofenced region and another. The path in the network is found in a shortest path fashion by means of a routing engine. Each trajectory is then constructed by starting at an initial timestamp and from the initial point of the path. By moving along the path a distance that is calculated based on a time increment $5s$ plus noise sampled from a Folded Gaussian distribution ($\mathcal{N}^{F}(0, 20)$) and a speed from a Gaussian distribution ($\mathcal{N}(4, 0.5)$) a new timestamp and a position can be determined. Finally, some noise is added to the longitudinal and latitudinal points of the position according to a Gaussian distribution $\mathcal{N}(0, 8.75)$. These values of the noise are chosen in order to simulate the inherent inaccuracy of a smartphone GPS sensor and to generate non-uniform spatial trajectories.

Anomalous trajectories are generated in the same way as the normal ones but the underlying path is enforced to visit one, two, three, or four manually selected intermediate points. 
An example of the trajectories resulting from this generation process restricted to a single pair of origin and destination regions is illustrated in Fig.~\ref{fig:intro}.

\subsection{Preprocessing of noisy data} \label{sub:noisy_data}

The synthetic trajectories were generated without extreme noise, but for trajectories in a real dataset extremely noisy geospatial points can be removed by checking whether they have an unrealistic high absolute acceleration. Geospatial points with an unrealistic high absolute acceleration (e.g., higher than the acceleration of an average car) can be identified as being noisy as it implies that there has been a large change in speed in a short amount of time. This makes it unlikely for the geospatial point to have been generated due to the natural movement of a person. The absolute acceleration $u_i$ associated with a geospatial point $\p_i$ can be calculated based on the change in speed and time between two consecutive geospaital points, that is, if the speed at point $\p_{i}$ is $v_{i} = \frac{d(\p_{i- 1}, \p_{i})}{\delta_{i} - \delta_{i - 1}}$ then the absolute acceleration at point $\p_{i}$ is $u_i = \frac{\left| v_{i} - v_{i - 1} \right|}{\delta_{i} - \delta_{i - 1}}$. This filtering approach can be incorporated into the preprocessing component \circled{A1} in Fig~\ref{fig:data_processing_flow}.

\subsection{Performance metrics \& statistics} \label{sub:performance_metrics}

To evaluate the performance of the detection methods in case-study (2) a couple of different performance metrics and statistics are calculated. Among others, the ROC Curve (Receiver Operating Characteristic Curve) is constructed to then be able to compute a corresponding AUC value that measures the performance of the detection methods.
The closer the AUC value is to $1$ the better performance the detection approach achieves.

The detection delay is another performance metric calculated to determine how fast an approach is able to detect a synthetically generated anomalous trajectory as actually being anomalous. The detection delay is calculated as the difference between the timestamp of the first point in an anomalous trajectory that can be said to deviate from a set of normal trajectories and the timestamp of a point in an anomalous trajectory where it has been detected by an approach as being anomalous. 

Beyond the AUC value and the detection delay, the median online and offline processing time is reported as well. The online processing time is the time it takes to process a single geospatial trajectory from start to finish to be able to determine if it is anomalous or not. The offline processing time on the other hand is the time it takes to prepare the historical data in the set $D$ to be used for real-time detection. For the proposed approach, the online processing time is essentially the time it takes to go through steps \circled{A1}--\circled{A6}, while the offline processing time is the time it takes to go through steps \circled{B1}--\circled{B6}.

\section{Experimental Results} \label{sec:experimental_results}

\subsection{Case Study 1} \label{sub:preclu}

To obtain the results mentioned in the following the preprocessing and segmentation steps were applied using parameter values $\tau = 600s$, $\epsilon = 300s$,  $\xi'=\xi'' = 28m$,  $\gamma = 500m$ and $\alpha = 100m$. These were determined based on their physical interpretation and relation described earlier in Section~{\ref{sec:preliminaries}} and through preliminary experiments.

\begin{figure}[t]
    \centering
    \includegraphics[trim={0.3cm 0.3cm 0.0cm 0.25cm},clip, width=0.50\textwidth]{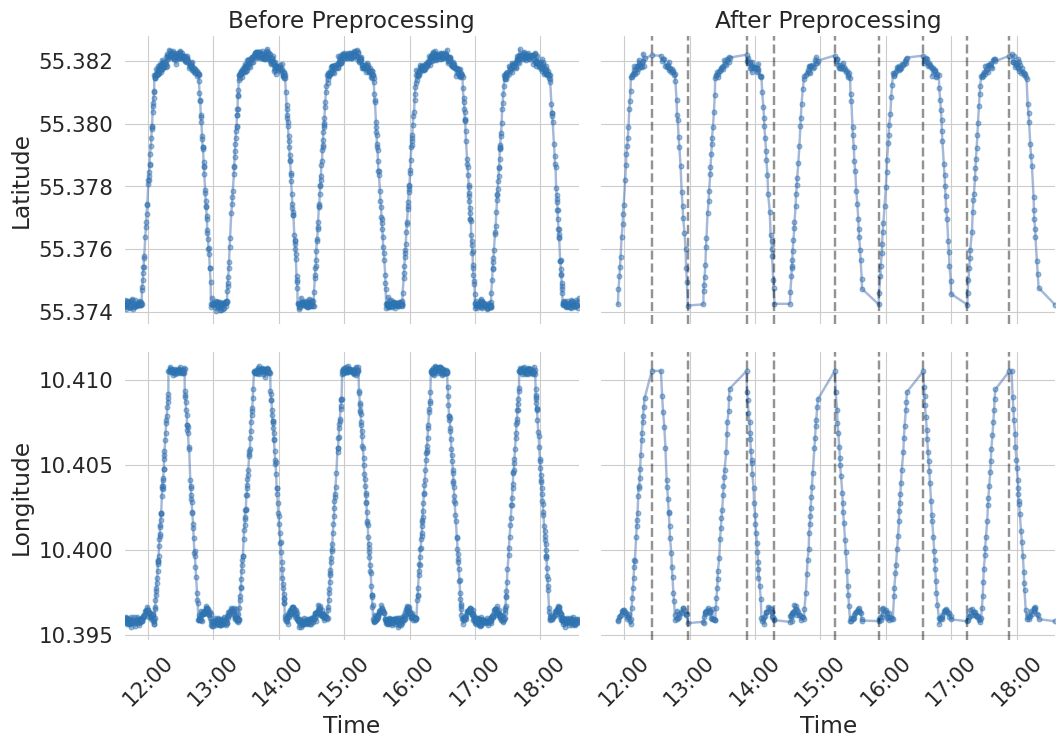}
    \caption{The lat/lon coordinates of a block of geospatial points plotted over time before and after preprocessing. The vertical lines indicate points at which a block is segmented and separate geospatial trajectories are created.}
    \label{fig:preprocess_result}
\end{figure}

The result of the preprocessing and segmentation steps is a set of compressed geospatial trajectories that goes between distinct geofenced regions. The unprocessed block of geospatial points is depicted in the first column of Fig.~\ref{fig:preprocess_result} in terms of its latitudinal and longitudinal coordinates. Each high and low peak essentially corresponds to an extended stay at a geofenced region. In the second column of Fig.~\ref{fig:preprocess_result}, the resulting geospatial trajectories are displayed.

The unprocessed block consists of 1209 geospatial points while the trajectories that result after the preprocessing has been applied consist of a total of 390 geospatial points, i.e., the number of geospatial points have been reduced by about $68\%$.

\subsection{Case Study 2} \label{sub:compdet}

We use the set of 222 trajectories in a leave-one-out validation model, in which each trajectory is used once as a test set for detection while the remaining trajectories are used as a training set to provide the historical data, similarly as in instance-based learning methods. For the iBDD method the historical trajectories are simply transformed into their corresponding sequence representation, while for the proposed detection approach the PrefixSpan algorithm is applied to mine frequent sequential patterns, which then represent the normal travel patterns of a person. 

For each possible train/test split the iBDD method and our detection approach are applied.  This yields a detection result (detected normal or anomalous) for each trajectory. Based on all the detection results, the AUC value can be calculated using the ground truth labels, which were associated with the trajectories when they were generated.

These steps are repeated for several different parameter settings:
\begin{itemize}
    \item For our approach we vary the geohash precision\footnote{\;The chosen geohash precision values correspond to maximum latitude/longitude encoding error in meters: $152.703$, $76.351$ and $38.176$, respectively.} $\left\{17, 18, 19\right\}$ and the anomaly threshold $\theta \in \left\{0.20, 0.40, 0.60\right\}$, while keeping the other parameters fixed: $\eta = 1$ along with $\tau = 600s$, $\epsilon = 300s$, $\xi'=\xi'' = 28m$,  $\gamma = 500m$ and $\alpha = 100m$.
    \item For the iBDD method we vary the geohash precision $\left\{17, 18, 19\right\}$ and support set threshold $\theta' \in \left\{0.20, 0.10, 0.05\right\}$.
\end{itemize}

We summarize the numerical results in Tables~\ref{tab:proposed} and \ref{tab:competing}.
From Table~\ref{tab:proposed} we observe that our proposed method generally performs well for most parameter values but especially for the highest threshold and geohash precision. Indeed the AUC value is high, close to 1, and the detection delay is small. However, the higher the geohash precision is the longer the fitting and detection times are. A low anomaly threshold ($\theta = 0.20$) results in faster anomaly detection but more false positives indicated by a worse AUC value.

From Table \ref{tab:competing} we see that the iBDD method generally performs worse for most parameter values. It is also possible to make some of the same observations w.r.t.\ the detection delay and geohash precision as with the results obtained from the proposed approach.

However, the iBDD method seems to be more efficient in terms of detection and fitting times in comparison with the proposed approach. This is mainly because it lacks the more sophisticated sequence mining and alignment components that the proposed approach has. These essentially improve the detection results at the cost of longer computation times.

In fact, assuming that each geospatial point $\p_i$ of a trajectory $T$ of size $|T|$ is mapped directly to a unique geohash value and a set of frequent sequential patterns $D$ is available where it holds that $\max_{\forall U \in D}(|U|) \leq |T|$, then the proposed detection approach runs in $\mathcal{O}(|D| \cdot |T|^2)$ time. This is primarily due to the repeated application of the Smith-Waterman algorithm in Eq.~{\eqref{eq:simcalc}}. On the other hand, if $D'$ is a set of historical geospatial trajectories that have been turned into geohash sequences, then the iBDD method runs in just $\mathcal{O}(|D'| \cdot |T|)$ time.

\begin{table}[h]
    \setlength\tabcolsep{4.0pt}
    \centering
    \caption{Case-study (2). Results from proposed method.}
    \label{tab:proposed}
    \begin{tabular}{llcccccc}
        \toprule
        \textbf{Threshold $\theta$} & \textbf{Precision}  & \multicolumn{2}{c}{\textbf{Statistics}} & \multicolumn{2}{c}{\textbf{Comp. time (s)}}       \\
            &   & AUC & Delay (s) & Detection & Fitting   \\
        \midrule
        0.20 & 17 & 0.7773 &  206.4440 & 0.0779 & \X\X4.7820 \\
             & 18 & 0.7793 &  165.1885 & 0.2661 &  \X11.0811 \\
             & 19 & 0.5765 & \X81.8410 & 1.2137 &   518.0548 \\

        \addlinespace[4.pt]
        0.40 & 17 & 0.9761 & 539.2180 & 0.0767 & \X\X4.7890 \\
             & 18 & 0.9926 & 366.2065 & 0.2644 &  \X11.0974 \\
             & 19 & 0.9687 & 325.5035 & 1.1856 &   516.6371 \\

        \addlinespace[4.pt]
        0.60 & 17 & 0.9564 & 534.7430 & 0.0774 & \X\X4.7981 \\
             & 18 & 0.9650 & 606.3950 & 0.2628 &  \X11.0942 \\
             & 19 & 0.9738 & 483.7400 & 1.1630 &   517.6821 \\
        \addlinespace
        \bottomrule
    \end{tabular}
\end{table}

\begin{table}[h]
    \setlength\tabcolsep{4.0pt}
    \centering
    \caption{Case-study (2). Results from the iBDD method.}
    \label{tab:competing}
    \begin{tabular}{llcccccc}
        \toprule
        \textbf{Threshold $\theta'$} & \textbf{Precision}  & \multicolumn{2}{c}{\textbf{Statistics}} & \multicolumn{2}{c}{\textbf{Comp. time (s)}} \\
            &   & AUC & Delay (s) & Detection & Fitting \\
        \midrule
        0.20 & 17 & 0.5311    & 269.1805  & 0.0327    & 3.6642  \\ 
             & 18 & 0.3117    & 107.1480  & 0.0448    & 4.0947  \\ 
             & 19 & 0.0025    & \X50.7300 & 0.0680    & 4.7236  \\

        \addlinespace[4.pt]
        0.10 & 17 & 0.6283    & 428.2890  & 0.0313    & 3.6557  \\ 
             & 18 & 0.4887    & 107.1480  & 0.0395    & 4.1346  \\
             & 19 & 0.0349    & \X37.6780 & 0.0546    & 4.6534  \\

        \addlinespace[4.pt]
        0.05 & 17 & 0.8624    & 462.5295  & 0.0318    & 3.6591  \\ 
             & 18 & 0.5288    & 135.4025  & 0.0384    & 4.1163  \\
             & 19 & 0.0624    & \X50.9360 & 0.0509    & 4.6573  \\
        \addlinespace
        \bottomrule
    \end{tabular}
\end{table}


\section{Conclusions and Future Work} \label{sec:conclusion}

We proposed an integrated framework for the online detection of wandering patterns of people with dementia. Our approach operates better than the state-of-the-art alternative, in terms of performance of detection and delay of detection.
We plan to devise and implement a more accurate detection solution that will be based on patterns extracted from raw trajectory data, rather than sequences of geohash values. By doing so we will be able to capture the movement of elderly people in more detail, both in the spatial aspect of their movement, as well as the temporal aspect (e.g., take into account speed or time of day).
Furthermore, we plan to go beyond the detection and propose a method for the prediction of these kinds of wandering patterns.
Finally, we plan to experiment with real data from the "Sammen Om Demens" project, a project with a local municipality, with a known ground truth that will help us draw more accurate conclusions.

\enlargethispage{-0.5in} 

\newpage
\bibliographystyle{IEEEtran}
\bibliography{IEEEabrv,bibliography}

\end{document}